\documentclass{article} 
\usepackage{graphicx}
\usepackage[ruled,linesnumbered]{algorithm2e}
\usepackage{subcaption}  
\usepackage{amssymb}
\usepackage{iclr2025_conference,times}
\usepackage{xcolor}
\usepackage{authblk}
\usepackage{booktabs}
\usepackage{colortbl}
\usepackage{amsmath}
\usepackage{booktabs} 
\usepackage{makecell}

\usepackage{amsmath,amsfonts,bm}

\def\eqref#1{equation~\ref{#1}}

\def\1{\bm{1}}

\DeclareMathAlphabet{\mathsfit}{\encodingdefault}{\sfdefault}{m}{sl}
\SetMathAlphabet{\mathsfit}{bold}{\encodingdefault}{\sfdefault}{bx}{n}

\usepackage{hyperref}
\usepackage{url}
\begin{document}

\title{Harnessing Diversity for Important Data Selection in Pretraining Large Language Models}

\makeatletter
\newcommand{\equalcontrib}{\textsuperscript{*}}
\newcommand{\coauthor}{\textsuperscript{\dag}}
\makeatother

\author[1]{Chi Zhang \equalcontrib}
\author[2]{Huaping Zhong \equalcontrib}
\author[1]{Kuan Zhang}
\author[1]{Chengliang Chai \coauthor}
\author[3]{Rui Wang}
\author[3]{Xinlin Zhuang}
\author[3]{Tianyi Bai}
\author[3]{Jiantao Qiu}
\author[4]{Lei Cao}
\author[5]{Ju Fan}
\author[1]{Ye Yuan}
\author[1]{Guoren Wang}
\author[3]{Conghui He \coauthor}
\affil[1]{Beijing Institute of Technology}
\affil[2]{SenseTime Research}
\affil[3]{Shanghai Artificial Intelligence Laboratory}
\affil[4]{University of Arizona/MIT}
\affil[5]{Renmin University of China}
\affil[ ]{\tt\small \{zc315, zhangkuan, ccl, yuan-ye, wanggrbit\}@bit.edu.cn \\ zhonghuaping@sensetime.com, \{heconghui, wangrui, zhuangxinlin, baitianyi, qiujiantao\}@pjlab.org.cn \\ lcao@csail.mit.edu \\ fanj@ruc.edu.cn}
\vspace{-0.5cm}
\renewcommand{\thefootnote}{\fnsymbol{footnote}}

%

\newcommand{\fix}{\marginpar{FIX}}
\newcommand{\new}{\marginpar{NEW}}
\newcommand{\lei}[1]{\textcolor{purple}{Lei: #1}}

\iclrfinalcopy 

\maketitle

\begin{abstract}
Data selection is of great significance in  pretraining large language models, given the  variation in quality within the large-scale available training corpora. 
%
To achieve this, researchers are currently investigating the use of data influence to measure the importance of data instances, $i.e.,$ a high influence score indicates that incorporating this instance to the training set is likely to enhance the model performance. Consequently, they select the top-$k$ instances with the highest scores.  
However, this approach has several limitations. 
(1) Calculating the accurate influence of all available data is time-consuming.
(2) The selected data instances are not diverse enough, which may hinder the pretrained model's ability to generalize effectively to various downstream tasks.
In this paper, we introduce \texttt{Quad}, a data selection approach that considers both quality and diversity by using data influence to achieve state-of-the-art pretraining results.
To compute the influence ($i.e.,$ the quality) more accurately and efficiently, we incorporate the attention layers to capture more semantic details, which can be accelerated through the Kronecker product. 
For the diversity, \texttt{Quad} clusters the dataset into similar data instances within each cluster and diverse instances across different clusters. For each cluster, if we opt to select data from it, we take some samples to evaluate the influence to prevent processing all instances. Overall, we favor clusters with highly influential instances (ensuring high quality) or clusters that have been selected less frequently (ensuring diversity), thereby well balancing between quality and diversity.  Experiments on Slimpajama demonstrate that ~\texttt{Quad} significantly outperforms other data selection methods with a low FLOPs consumption. Further analysis also validates the effectiveness of our influence calculation. Our code and data are available at (\url{https://anonymous.4open.science/r/Quad/}).
\end{abstract}



\section{Introduction}
Recently, large language models (LLMs) have significantly advanced the field of artificial intelligence~\citep{zhao2023survey, hadi2023survey, minaee2024large}. Due to the unprecedented number of parameters (model size) and the pre-training on huge amount of training data, LLMs are generalizable a broad spectrum of downstream tasks.
However, in practice, the computation resources limit both the model size and the volume of data used in pre-training.  In this situation, judiciously selecting train datasets is critical for producing highly performance LLMs~\citep{brown2020language, du2022glam, gururangan2020don, hoffmann2022empirical, raffel2020exploring}. In particular, the quality of the training datasets vary dramatically, while the LLaMA-3.1 report~\citep{dubey2024llama} shows that the use of high quality data in later training stages can greatly improve model performance.



Typical straightforward data selection approaches include rule-based data filtering~\citep{raffel2020exploring, rae2021scaling}, querying high-performance models (\textit{e.g.}, GPT-4)~\citep{wettig2024qurating, sachdeva2024train}, surrogate models~\citep{lin2024rho, shao2024deepseekmath}, etc. Although these methods have achieved success on some datasets and models, they rely on simple heuristics to select training data. Without explicitly measuring the impact of the selected data on the model, these methods tend to produce sub-optimal pretraining results.  
To address this issue, some researchers~\citep{xialess, yu2024mates} start evaluating each data instance by assigning it a score that reflects its impact on the model. Frequently used scoring methods include the influence function~\citep{xialess}, early loss~\citep{albalak2023efficient}, and perplexity~\citep{chen2024towards}. Among these methods, the influence function consistently delivers state-of-the-art results by effectively approximating the impact of adding each instance to the training set.
A higher score signifies a higher priority for selecting a data instance, and hence the top-$k$ (or gumble top-$k$) instances with the highest scores are chosen~\citep{xie2023data, wettig2024qurating, yu2024mates}.


However, the above methodologies have the following limitations. 

\noindent\textbf{Prohibitive Computation Cost.}
First, accurately calculating the influence score of one data instance is expensive, because it involves the computation of the Hessian matrix. However, in the LLM pre-training, the number of the candidate data instances is extremely large. It is thus prohibitively expensive to compute the scores for all of the candidates.

\begin{figure}
  \begin{minipage}[h]{0.5\linewidth}
    \centering
    \includegraphics[scale=0.091]{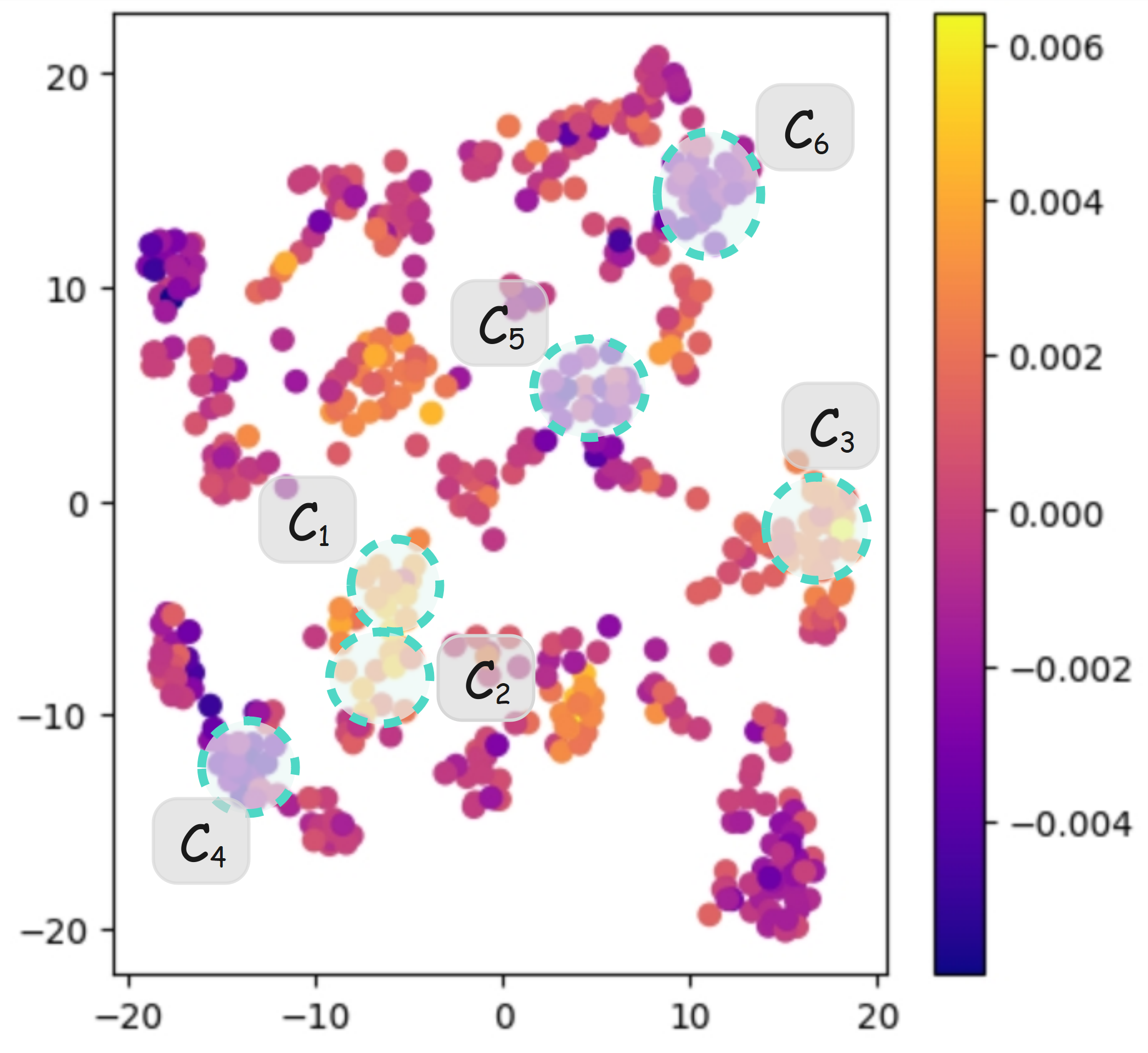}
    \subcaption{Influence scores of data instances}
    \label{fig1}
  \end{minipage}%
  \begin{minipage}[h]{0.5\linewidth}
    \centering
    \includegraphics[scale=0.091]{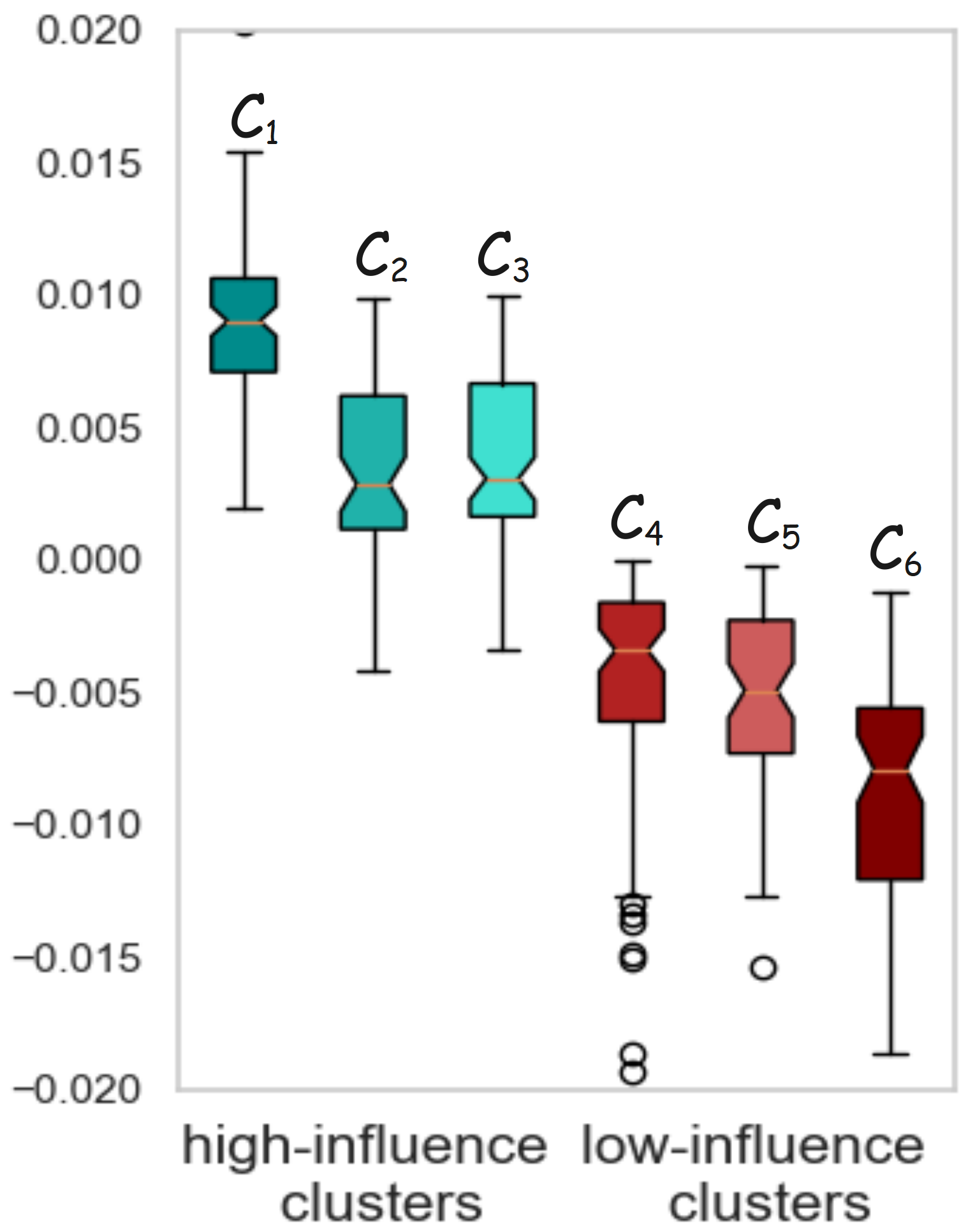}
    \subcaption{Influence scores in different clusters}
    \label{fig2}
  \end{minipage}
  \caption{Distribution of influence scores of some sampled data instances.}
  \label{fig:cluster_analysis}
\end{figure}

\noindent\textbf{Lack of Diversity.}
Second, assume that all influence scores have been calculated, as shown in Figure~\ref{fig1}.  We can see that the top-$k$ instances ($e.g.,$ some high-score instances in $C_1$) tend to be closely distributed in the feature space because the influence computation is closely related to the data features. 
That is, the training instances selected in this way are lack of diversity ($e.g.,$ other instances in $C_3$ with high influence are also worth selecting), while as confirmed by some studies~\citep{abbas2023semdedup, tirumala2023d4}, diversifying training samples mitigates overfitting, thereby enhancing the generalizability of the model.
Therefore, an effective data training selection method should take both the influence scores and the diversity into consideration.


We thus propose \texttt{Quad}, a scalabe and effective data selection approach, which successfully addressing above challenges, achieves state-of-the-art pretraining results. Initially, \texttt{Quad} organizes the given dataset into clusters where the data instances within each cluster are similar, and those in different clusters exhibit diversity. Hence, we can sample a data subset from a cluster to estimate the accurate average influence of the cluster, so as to represent the cluster quality $w.r.t$ the model performance. 

Next, leveraging the property of the attention-based Transformer architecture which is widely adopted by the LLMs, we design a novel method to accurately compute the influence of an instance on LLM pre-training. More specifically, rather than solely relying on the MLP layers to compute the influence~\citep{koh2017understanding, yu2024mates, grosse2023studying, engstrom2024dsdm}, we incorporate the attention layers such that the influence computation considers more semantic information. In addition, given that calculating the Hessian matrix is time-consuming, particularly for attention layers with complex interactions, we incorporate the Kronecker product to approximate  the Hessian matrix, thereby greatly expediting the computation. This successfully addresses the computation cost challenge.

To improve diversity, we apply the Multi-Arm Bandit (MAB) technique, where each cluster is regarded as an arm of the MAB. Upon selecting an arm, we draw samples from the cluster to calculate influence scores. Subsequently, \texttt{Quad} iteratively samples from clusters, taking into account both the influence score and data diversity, e.g., whether the cluster has already been sampled. Moreover, because this sampling strategy effectively avoids calculating the influence of all instances, it further speeds up the data selection process.



We summarize our main contributions as follows:

\begin{itemize}

\item To balance the quality and diversity, we incorporate an iterative MAB solution to first cluster the data instances and select data instances from these clusters.

\item We propose a novel method to compute the influence function in attention-based Transformer architecture, so as to precisely measure the data quality in LLM pre-training.

\item Experiments on the widely-used dataset Slimpajama and 9 popular downstream tasks demonstrate that ~\texttt{Quad} significantly outperforms state-of-art data selection methods by 1.39\% in zero-shot accuracy, also with low computation resources consumption. 

\end{itemize}

\

\section{Related Work}



\textbf{Rule-based Methods.}
Initially, researchers often relied on intuition to design hand-crafted heuristics~\citep{soldaini2024dolma} and ~\citep{penedo2023refinedweb}, aiming to improve data quality. Deduplication is another typical approach for selecting pretraining data, such as ~\citep{penedo2023refinedweb} and SemDedup ~\citep{abbas2023semdedup} which use keyword-based and semantic deduplication, respectively. Additionally, certain approaches employ $n$-gram similarity~\citep{gao2020pile, xie2023data} to assist in choosing corpora that is semantically aligned with the validation set data.
Although these methods effectively filter out noise and redundant data from web sources, they rely on simple heuristics and cannot be well generalized.

\textbf{LLM As a Selector.}
Although large models such as GPT-4 can effectively assess data quality due to their semantic comprehension capacity, the metrics utilized to rate data ($e.g.,$ writing style, educational value etc.) heavily rely on human intuition~\citep{wettig2024qurating, penedo2024fineweb, zhang2024autonomous, gunasekar2023textbooks}. This often leads to a mismatch between the selected data and the data desired by the model. 


\textbf{Surrogate Models.}
DeepSeekMath ~\citep{shao2024deepseekmath} proposes an active learning strategy to train a web data classifier. Similarly, in MATES~\citep{yu2024mates}, a surrogate model was developed to estimate the influence scores of the data instances. RHO-1 ~\citep{lin2024rho} used a surrogate model trained with high-quality data to perform token-level data filtering. 
However, these surrogate models are not trained over large-scale data, and thus their generalizatio ability is limited.




\textbf{Perplexity} serves as a metric for selecting high-probability data in a language model. In ~\citep{chen2024towards, marion2023less, muennighoff2024scaling, wenzek2019ccnet}, perplexity (PPL) is utilized to filter data. As also discussed in Qurating~\citep{wettig2024qurating}, we observe that this method often incorporates a significant amount of simple and redundant data, because they are easy for the model to predict.

\textbf{Influence Function} 
~\citep{grosse2023studying, choe2024your} demonstrates that influence function can reveal the impact of training data on the performance of large models. Consequently, LESS~\citep{xialess} and MATES ~\citep{yu2024mates} utilize influence functions for selecting data during the SFT and pretraining phases, respectively. For large models, computing influence functions is computationally expensive. ~\citep{grosse2023studying}. Hence, given the large amount of data handled during pretraining, directly using LESS~\citep{xialess} for data selection at this stage poses considerable difficulties. To overcome this, MATES~\citep{yu2024mates} employs a proxy model to approximate the influence score across the full dataset. 
However, the limited capacity of this small proxy model hinders its ability to provide accurate influence scores.
Furthermore, relying on the influence to select data solely often leads to a lack of diversity in the chosen data. 

\section{Methods}
First, we present our problem statement in \S \ref{sec:framwork}. Next, in \S \ref{sec:mab}, we explain how our method achieves the  balance between quality and diversity in selecting pretraining data. Finally, in \S \ref{sec:ek-fac}, we 
introduce how we compute the influence with attention layers more accurately and efficiently.

\begin{figure}[h]
\begin{center}
\includegraphics[width=1\textwidth]{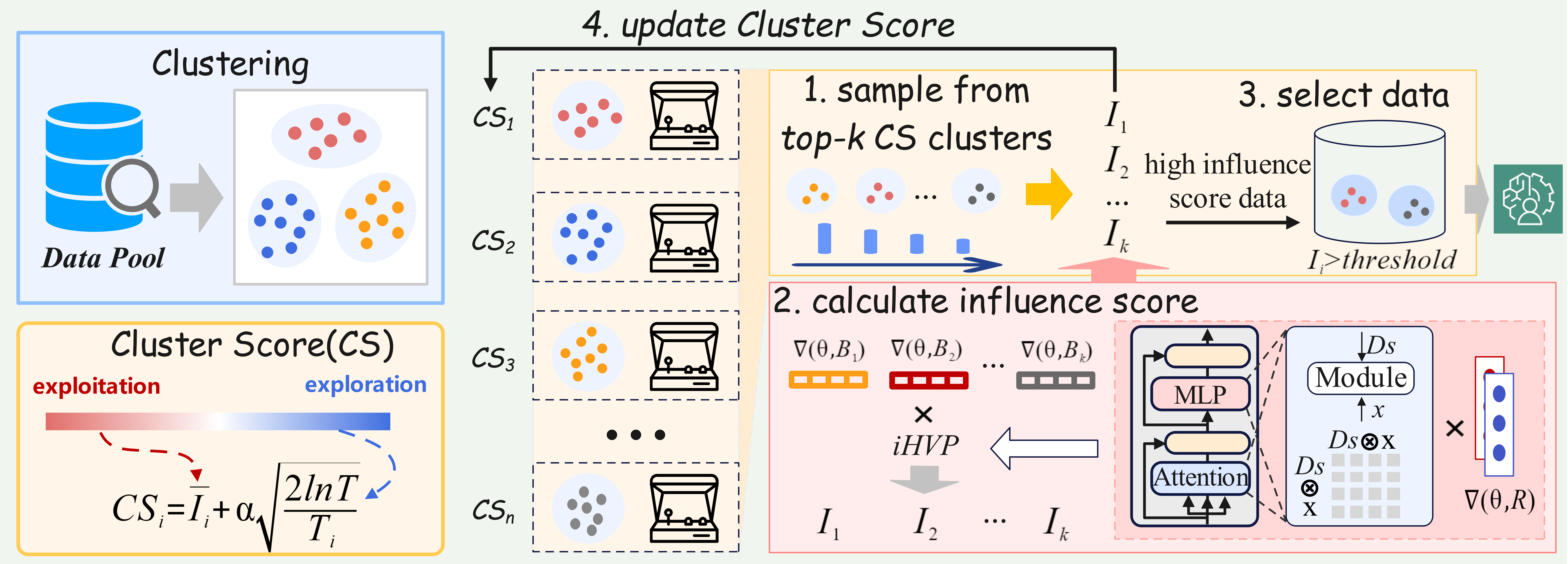}
\end{center}
\caption{Overview of \texttt{Quad}}
\label{fig3}
  
  \begin{minipage}{1\textwidth}
    \small  
  \end{minipage}
\end{figure}



\subsection{Problem Definition}
\label{sec:framwork}
To enhance the capabilities of large models, it is necessary to retrieve relevant data from a large pool of candidate data and perform further training for the large model. 
Formally, given the pool $D_c$ and a reference set $D_r$, our problem is to select a subset  $D_b \subset D_c$  to fine-tune the large model $M$, with the aim of minimizing the loss of the updated model $M'$ in the reference set $D_r$.



\subsection{balance between quality and diversity}~\label{sec:mab}
As shown in Figure~\ref{fig2}, there are significant variations in the distribution of influence scores among different clusters.
To achieve the quality-diversity balance, it is necessary to know the precise average influence score for instances in each cluster.
However, Figure~\ref{fig2} shows that the influence scores for each cluster also fluctuate around the average, indicating a certain level of uncertainty. Estimating the average with a small sample size will not be accurate enough, while taking a large number of samples to compute the average influence is costly.
%

Hence, we propose to use the MAB~\citep{vermorel2005multi} technique that is capable of making decisions iteratively under uncertainty. At a high level, each cluster represents an arm of the MAB, and during each iteration, a cluster with a high average influence score tends to be selected and sampled. We will then compute the influence of data instances to update the average. Moreover, clusters that are not visited often present significant opportunities for sampling to balance the diversity.



The overall process of this approach is illustrated in Figure~\ref{fig3}. Specifically, our method can be divided into the following four steps: First, we \textit{sample the top-$k$ clusters} with the highest cluster scores (denoted by CS) computed by MAB. 
Here, the cluster score is determined by both the influence score and the sample frequency.
Then we \textit{calculate the influence scores} for the samples in each cluster (Section 3.3). At this point, we \textit{select  high scoring samples} to be added for training and use their scores to \textit{update the cluster score} for each cluster.
Throughout the iterative process, the MAB algorithm focuses on frequently sampling high-quality clusters that have high influence scores, which also  enhances the accuracy of their quality estimation ($i.e.,$ updating the average influence $\overline{I}_i$). Simultaneously, it ensures diversity by also sampling less-visited clusters.
Next, we discuss how to compute and update the cluster score in details.





\textbf{Cluster Score (CS).} The Upper Confidence Bound can effectively balance exploration ($i.e.,$ data diversity) and exploitation ($i.e.,$ data quality), so we use it as the cluster score to evaluate each cluster, as shown in Equation (1). 
Specifically, the cluster score is determined by the average influence score $\bar{I_i}$ and the exploration score $\sqrt{\frac{2\ln {\sum_j{T(C_j)}}}{T(C_i)}}$, where $T(C_i)$ denotes the frequency of instances sampled from cluster $C_i$, and $\sum_j T(C_j)$ denotes  the total times of samples taken from all clusters.

\begin{equation}
    CS_{i}=\bar{I_i}+\alpha\sqrt{\frac{2\ln {\sum_j{T(C_j)}}}{T(C_i)}}
\end{equation}

\textbf{Update the cluster score.} During each iteration, a subset of data $B_i$ is sampled from each cluster with a high cluster score (CS). 
The sum of their influence score $I_i$ can be used to denote the impact of the samples from the cluster $C_i$ on the model.



\begin{equation}
    R(C_i) += \sum_{z \in B_i}I_\theta(D_r,z)  , \quad T(C_i) += 1
\end{equation}

where $R(C_j)$ denotes the total reward accumulated by cluster $C_i$ over several iterations.
Then the average influence score $\bar{I_{i}}$ for cluster $C_i$ can be represented as
    $\bar{I_i} = \frac{R(C_i)}{T(C_i)}$.
As the sample size grows, $\bar{I_i}$ for each cluster $C_i$ steadily approaches the exact average influence of the cluster, which can be used to update the cluster scores for all clusters.

\textbf{Data selection.} During each iteration, we pick a small proportion($\gamma$) of data instances from selected clusters. We also require that these instances have influence scores higher than the threshold $\tau$, otherwise we will not select them, which are then added into the training dataset.



\begin{algorithm}[h]
    \caption{\texttt{Quad} Algorithm}
    \KwIn{Candidate data pool $D_c$, reference set $D_r$, the model $\theta$}
    \KwOut{Selected data $D_b$}
    $\mathcal{C}$ = Cluster($D_c$)\;
    \While{}{
        $C_{top\_k}$ = top-$k$ clusters with the highest Cluster Score(CS) \;

        $B_{top\_k}$ = mini-batchs sampled from $C_{top\_k}$
            
        \For{$C_i$ in $C_{top\_k}$}{      
                \quad $R(C_j)$ += $\sum_{z \in B_i}I_\theta(D_r,z),$\quad $T$($C_j$) += 1 \;                
        }
        \For{ $C_i$ in $\mathcal{C}$}{
            $\bar{I_i} = \frac{R(C_i)}{T(C_i)}$ \;
            \textbf{if} $\bar{I_i} > threshold$ \textbf{then} $D_b += \gamma C_i$\;
        }
        $CS_{i}=\bar{I_i}+\alpha\sqrt{\frac{2\ln {\sum_j{T(C_j)}}}{T(C_i)}}$ \;
    }
    \Return{$D_b$}\;
\end{algorithm}
\subsection{Influence Calculation with attention layers}
\label{sec:ek-fac}
Instead of retraining the large model with each data sample $z$, the impact of  $z$ on the model $M$ can be estimated by calculating the influence function for each instance. In this section, we extend the influence calculation to multi-head attention layers and provide acceleration techniques.

\begin{equation}
    I_\theta(D_r,z)=-\nabla L(\theta,D_r)(H + \lambda I)^{-1}\nabla L(\theta,z)
\end{equation}


%


In the above equation, $I_\theta(D_r,z)$ denote the influence function of data $z$ on model $\theta$. $\nabla L(\theta,D_r)$ and $\nabla L(\theta,z)$ denote the gradient of reference dataset $D_r$ and data $z$, respectively. Since the training of the large model  does not often fully converge, resulting in a non-invertible Hessian matrix $H$, a regularization term $\lambda I$ is introduced~\citep{bae2022if}. Equation (3) is typically divided into the following two stages to speed up the computation:

1. Approximate the multiplication of the gradient of the validation set $\nabla L(\theta,D_r)$ and the inverse Hessian matrix $H^{-1}$  using the inverse Hessian vector product (iHVP).

2. Compute the dot product between the iHVP and the gradient of each training data point $\nabla L(\theta,z)$.

While this framework can accelerate the computation of the influence function, scaling it up to large language models (LLMs) with massive parameters is still expensive. Hence, K-FAC~\citep{martens2015optimizing, ueno2020rich} can be used to accelerate the iHVP computation by using the Kronecker product to decompose the Hessian matrix.

The K-FAC approximate the parameters of different MLP layer $\theta_1$, $\theta_2$ and $\theta_3$ as independent. That's because, during the gradient computation and update process, there are usually only minimal direct dependencies between the gradients of different MLP layers. This is particularly evident during back propagation, where the weight updates for each MLP layer are primarily influenced by the parameters of that specific layer. Therefore, the influence function $I_{\theta_1, \theta_2, \theta_3}(D_r, z)$ in K-FAC method can be expressed as:

\begin{equation}
    I_{\theta_1, \theta_2, \theta_3}(D_r, z) = I_{\theta_1}(D_r, z) + I_{\theta_2}(D_r, z) + I_{\theta_3}(D_r, z)
\end{equation}

\begin{figure}[h]
\begin{center}
\includegraphics[width=1\textwidth]{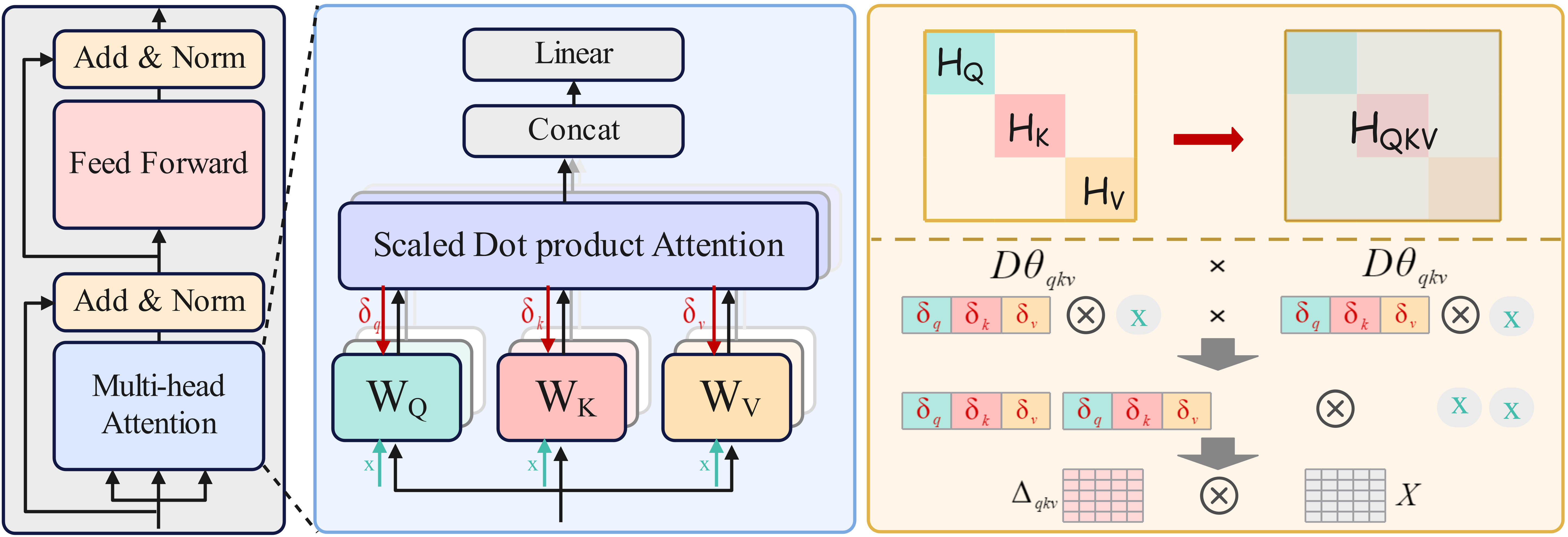}
\end{center}
\caption{Kronecker Product in calculating iHVP}
\label{multi-head-attention}
\end{figure}

In attention mechanisms, there exist complex connections between the Query, Key, and Value layers. As the right-upper corner of Figure~\ref{multi-head-attention} shows, separately calculate the hessian matrix of Query, Key and Value layers, will miss massive information of 
Consequently, it is essential to consider the QKV layers as a unified layer $\theta_{qkv}$ when computing the influence function. Therefore, the influence function $I_{\theta_{att}}(D_r, z)$ can be expressed as:
\begin{equation}
I_{\theta_{att}}(D_r, z) = I_{\theta_{qkv}}(D_r, z) + I_{\theta_{o}}(D_r, z)
\end{equation}


Then, as the right-lower corner of Figure~\ref{multi-head-attention} shows, by decomposing the Hessian matrix into a kronecker product of smaller matrices and computing the inverse of each smaller matrix, we can avoid directly inverting the entire Hessian matrix, significantly reducing computational cost, and accelerate this process:

\textbf{Forward propagation}:
\begin{equation}
    Attention(Q,K,V) = softmax(\frac{QK^{T}}{\sqrt{d_k}})V
\end{equation}


\textbf{Backward propagation}:
\begin{equation}
    D\theta =vec(DW)=\delta\otimes x
\end{equation}
Here, $\otimes$ denotes the Kronecker product, and $vec()$ represents the vectorization operation. Thus, the gradient of $\theta_{qkv}$ can be written as:

\begin{equation}
    D\theta_{qkv} = \begin{bmatrix} \mathbf{vec(DW_Q)} \\ \mathbf{vec(DW_K)} \\ \mathbf{vec(DW_V)} \end{bmatrix} \\
 = \begin{bmatrix} \mathbf{\delta}_q \\ \mathbf{\delta}_k \\ \mathbf{\delta}_v \end{bmatrix} \otimes x
\end{equation}
Let $\delta_{qkv}$
 $= \begin{bmatrix} \mathbf{\delta}_q \\ \mathbf{\delta}_k \\ \mathbf{\delta}_v \end{bmatrix} $. Then, the Hessian matrix $H_{qkv}$ can be estimates by:
\begin{equation}
\begin{aligned}
    H_{qkv} = E(D\theta_{qkv} {D\theta_{qkv}}^T)=E(\delta_{qkv}\delta_{qkv}^T\otimes x_{qkv}x_{qkv}^T) \\
    \approx E(\delta_{qkv}\delta_{qkv}^T) \otimes E(x_{qkv}x_{qkv}^T) = \Delta_{qkv} \otimes X_{qkv}
\end{aligned}
\end{equation}

Also, $H_o = \Delta_o \otimes X_o$. Thus, the iHVP of the attention layer can be estimated as follows: 

\begin{equation}
\begin{aligned}
H_{att}^{-1}v_{att} = 
\begin{bmatrix}\mathbf {H_{qkv}^{-1}v_{qkv}} \\ \mathbf{H_o^{-1}v_o}  \end{bmatrix}
= \begin{bmatrix} \mathbf{(\Delta_{qkv} \otimes X_{qkv})^{-1}v_{qkv}} \\ \mathbf{(\Delta_{o} \otimes X_o)^{-1}v_o} \end{bmatrix} \\
= \begin{bmatrix} \mathbf{(\Delta_{qkv}^{-1} \otimes X_{qkv}^{-1})v_{qkv}} \\ \mathbf{(\Delta_{o}^{-1} \otimes X_o^{-1})v_o} \end{bmatrix}
= \begin{bmatrix} \mathbf{vec(\Delta_{qkv}^{-1} V_{qkv} X_{qkv}^{-1})} \\ \mathbf{vec(\Delta_{o}^{-1} V_o X_o^{-1})} \end{bmatrix}
\end{aligned}
\end{equation}

where $v_{att}$, $v_{qkv}$, $v_{o}$ represent the gradient of reference dataset $D_r$ on parameters $\theta_{att}$, $\theta_{qkv}$, $\theta_{o}$, respectively. Thus, the influence score of attention layers can be written as:  $I_{\theta_{att}} = -\nabla L(\theta_{att}, z)H_{att}^{-1}v_{att}$.

To avoid the excessive memory usage of validation set gradients, we apply the Johnson-Lindenstrauss Lemma to reduce the dimensionality of both the iHVP computation results and the training data gradients $\nabla L(\theta,z)$.
\section{Experiment}

\subsection{Experiment Setup}

\textbf{Dataset Preparation.}
We use the entire 627B-token SlimPajama dataset as the candidate pool $D_c$. In the clustering process, the BAAI/bge-large-en-v1.5 model is employed to generate embeddings for the input data, and approximately 600 million data points from the candidate pool $D_c$ are clustered into 10,000 groups using the $k$-means algorithm. We use LAMBADA ~\citep{paperno2016lambada} as our reference set $D_r$, which is a widely used language modeling task and often serves as a validation benchmark for language model pretraining. ~\citep{yu2024mates, xie2023data, hoffmann2022empirical}.


\textbf{Experimental settings.}
We train a transformer-based decoder-only language model that contains 1.3B parameters, uses RoPE embeddings ~\citep{su2023enhanced}, and has a maximum context window of 1024 tokens ~\citep{touvron2023llama}. 
%
Following the setting of MATES~\citep{su2023enhanced}, 30B tokens out of the 627B are selected for training using \texttt{Quad} and compare with baselines.
%
%
The learning rate is set to $5 \times 10^{-5}$, the batch size is set to 4096, and the Adam optimizer is employed with hyperparameters $\beta_1 = 0.9, \beta_2 = 0.95, \epsilon = 10^{-8}$. As for Multi-Armed Badit, we set the $\alpha$ = 0.002 , sample proporation $\gamma$ = 0.05 and the sample threshold $\tau$ as 0.0025.

\textbf{Baselines.} We compare our methods with several baselines. (1) \texttt{Random} samples  data from the entire candidate dataset randomly. (2) \texttt{Qurating} uses the large language model to select data. (3) \texttt{DSIR} selects data instances that are similar to the LAMBADA dataset. (4) \texttt{PPL} uses perplexity-based data selection, $i.e.,$ selecting data instances with the lowest perplexity scores. (5) \texttt{MATES} trains a surrogate model to evaluate the influence of each data instance on the target model.

\textbf{Evaluation datasets.}
To comprehensively evaluate the capabilities of pretrained models, we conduct  experiments on various downstream tasks covering three significant categories:

General Knowledge: ARC-C, ARC-E~\citep{clark2018think}, and SciQ ~\citep{welbl2017crowdsourcing}.

Commonsense Reasoning: HellaSwag ~\citep{zellers2019hellaswag}, SIQA ~\citep{sap2019social}, WinoGrande~\citep{sakaguchi2021winogrande}, Logiqa ~\citep{liu2020logiqa}.

Reading Comprehension: OpenbookQA ~\citep{mihaylov2018can}, and BoolQ~\citep{clark2019boolq}.

Evaluations are conducted using the lm-evaluation-harness~\citep{gao10256836framework} framework  and the average accuracy ($i.e.,$ Overall Score) is reported for comparison.


\subsection{Results}

\textbf{Overall Performance.}
As demonstrated in Table~\ref{table1}, our method surpasses all the baseline methods in downstream tasks with zero-shot evaluation. 
To be specific, we can observe that on General Knowledge and Reading Comprehension tasks, \texttt{Quad} has the improvement of 1.75\% and 1.98\% respectively compared with \texttt{Random}.
\texttt{Quad} outperforms \texttt{DSIR} and \texttt{Semdedup} because they use rule-based heuristics to select data without considering the model. Although \texttt{PPL} and \texttt{MATES} consider the model, they do not perform well because the former one always selects some simple and duplicated instances, and the surrogate model of the latter one is small and lacking of enough training data.  \texttt{Qurating} generally performs the best among other baselines, but still worse than our approach, and it incorporates the highest FLOPS(1e19) because of the usage of LLMs for data selection.
 In terms of the FLOPs, we can observe that except the methods ($i.e.,$ \texttt{DSIR}, \texttt{SemDeDup}) that use simple heuristics, we consume minimal computation resources because our MAB solution samples from clusters without considering the entire candidate dataset like \texttt{PPL}, \texttt{Qurating} and \texttt{MATES}.
 


\renewcommand{\arraystretch}{1.05} 
\begin{table}[t]
\caption{Overall Performance}
\begin{center}
\scalebox{0.9}{
\begin{tabular}{cccccc}
\toprule
\textbf{Selection Method} & \makecell{\textbf{General}\\ \textbf{Knowledge}\\ \textit{(3 tasks)}} & \makecell{\textbf{Commonsense}\\ \textbf{Reasoning}\\ \textit{(4 tasks)}} & \makecell{\textbf{Reading}\\ \textbf{Comprehension}\\ \textit{(2 tasks)}} & \textbf{Overall} & \textbf{FLOPs}  \\ \midrule
Random                                                        &          50.33         &           36.19            &            39.09           &    41.55  &7.66   \\ \midrule
DSIR                                               &        50.37\textsuperscript{\colorbox{green!15}{\textcolor{black}{$\uparrow$0.04}}}           &             34.01\textsuperscript{\colorbox{red!15}{\textcolor{black}{$\downarrow$2.18}}}          &             38.80\textsuperscript{\colorbox{red!15}{\textcolor{black}{$\downarrow$1.29}}}          &    40.53\textsuperscript{\colorbox{red!15}{\textcolor{black}{$\downarrow$1.02}}}    &7.66\\ 

PPL                                                          &          48.71\textsuperscript{\colorbox{red!15}{\textcolor{black}{$\downarrow$1.62}}}         &            \textbf{37.72\textsuperscript{\colorbox{green!15}{\textcolor{black}{$\uparrow$1.53}}}}           &            38.57\textsuperscript{\colorbox{red!15}{\textcolor{black}{$\downarrow$0.52}}}           &     41.57\textsuperscript{\colorbox{green!15}{\textcolor{black}{$\uparrow$0.02}}}  & 9.51  \\ 

Semdedup                                                         &         50.99\textsuperscript{\colorbox{green!15}{\textcolor{black}{$\uparrow$0.66}}}          &            36.11\textsuperscript{\colorbox{red!15}{\textcolor{black}{$\downarrow$0.08}}}           &            39.44\textsuperscript{\colorbox{green!15}{\textcolor{black}{$\uparrow$0.35}}}          &     41.81\textsuperscript{\colorbox{green!15}{\textcolor{black}{$\uparrow$0.26}}} & 8.11   \\ 

Qurating                                                      &           \underline{51.56}\textsuperscript{\colorbox{green!15}{\textcolor{black}{$\uparrow$1.23}}}        &           35.93\textsuperscript{\colorbox{red!15}{\textcolor{black}{$\downarrow$0.26}}}            &             39.70\textsuperscript{\colorbox{green!15}{\textcolor{black}{$\uparrow$0.61}}}          &    \underline{42.01}\textsuperscript{\colorbox{green!15}{\textcolor{black}{$\uparrow$0.46}}} &13.66   \\ 

MATES                                                         &     50.45\textsuperscript{\colorbox{green!15}{\textcolor{black}{$\uparrow$0.12}}}              &       36.06\textsuperscript{\colorbox{red!15}{\textcolor{black}{$\downarrow$0.13}}}                &      \underline{39.83}\textsuperscript{\colorbox{green!15}{\textcolor{black}{$\uparrow$0.74}}}                 &   41.93\textsuperscript{\colorbox{green!15}{\textcolor{black}{$\uparrow$0.38}}}    &9.81  \\ \midrule

Quad(ours)                                                          &         \textbf{52.08\textsuperscript{\colorbox{green!15}{\textcolor{black}{$\uparrow$1.75}}}}          &           \underline{37.03}\textsuperscript{\colorbox{green!15}{\textcolor{black}{$\uparrow$0.84}}}            &           \textbf{41.07\textsuperscript{\colorbox{green!15}{\textcolor{black}{$\uparrow$1.98}}}}            &     \textbf{42.94\textsuperscript{\colorbox{green!15}{\textcolor{black}{$\uparrow$1.39}}}} &9.15 \\ \bottomrule
\end{tabular}}

\end{center}
\label{table1}
\end{table}

\begin{figure}
  \begin{minipage}[t]{0.25\linewidth}
    \centering
    \includegraphics[scale=0.23]{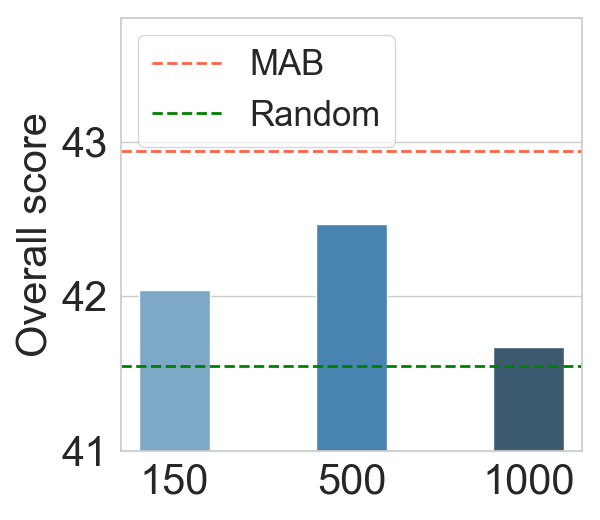}
    \subcaption{MAB vs. top-$k$ clusters}
    \label{experiment-fig1}
  \end{minipage}%
  \begin{minipage}[t]{0.25\linewidth}
    \centering
    \includegraphics[scale=0.23]{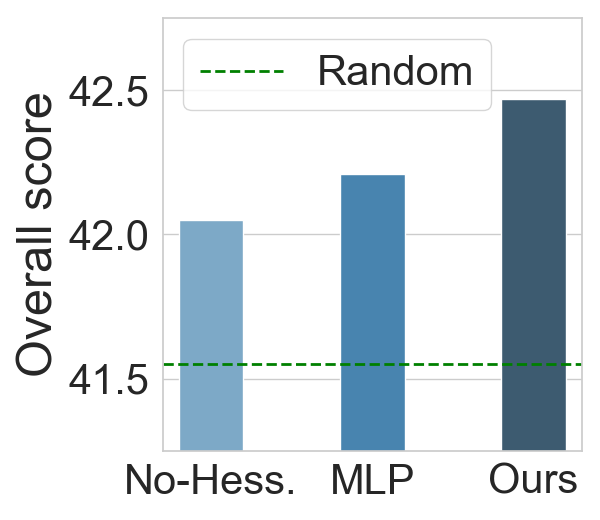}
    \subcaption{Influence accuracy}
    \label{experiment-fig2}
  \end{minipage}%
  \begin{minipage}[t]{0.25\linewidth}
    \centering
    \includegraphics[scale=0.23]{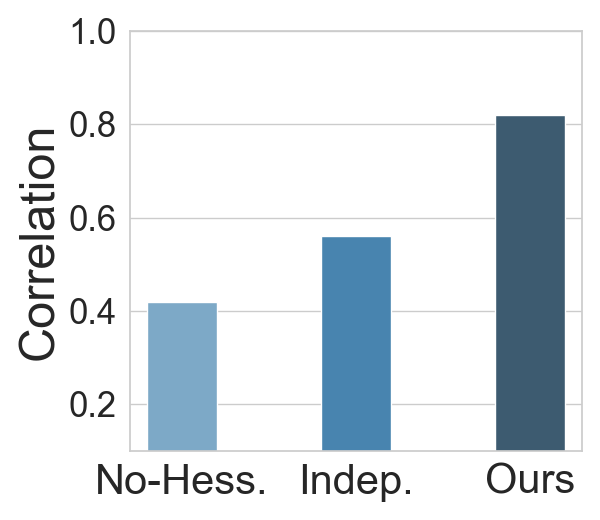}
    \subcaption{Influence correlation}
    \label{experiment-fig6}
  \end{minipage}%
  \begin{minipage}[t]{0.25\linewidth}
    \centering
    \includegraphics[scale=0.23]{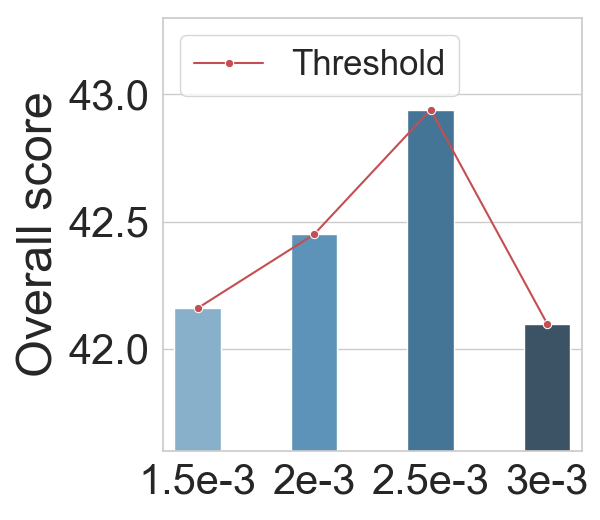}
    \subcaption{Sample threshold $\tau$}
    \label{experiment-fig4}
  \end{minipage}
  \caption{(a) shows the effectiveness of the MAB method; (b) shows the accuracy of calculating the influence function on MLP and attention layers; (c) shows the correlation between Query, Key, Value layers impact a lot on the accuracy of influence calculation; (d) shows the model performance of varying sample threshold $\tau$.}
  \label{fig:experiment2}
\end{figure}

\textbf{Effectiveness of MAB.}
This section evaluates the effectiveness of the MAB approach for data selection in contrast to the straightforward method of choosing the top-$k$ clusters with the highest influence scores for model training. To be specific, we randomly select an equivalent number of data points from the top 150, 500, and 1000 clusters.  Figure~\ref{experiment-fig1} illustrates the trade-off between data quality and diversity: clusters with higher influence scores do not necessarily enhance model performance on downstream evaluation sets because of their lack of diversity. Hence, the multi-armed bandit method can more effectively capture the trade-off between quality and diversity across clusters, resulting in  superior performance on downstream evaluation sets, as opposed to merely choosing the top-$k$ clusters.




\textbf{Effectiveness of Influence Calculation.}
This experiment studies the effectiveness of our influence calculation method. In this section, we select the top 500 clusters with the highest scores using three methods: (1) \texttt{no-Hessian} ($i.e.,$ computing the gradient similarity between  training data and reference data~\citep{pruthi2020estimating}) without considering the Hessian matrix; (2)  \texttt{MLP}($i.e.,$ calculating influence function on MLP layers) and (3) \texttt{Ours} ($i.e.,$ calculating influence function on both MLP and attention layers).
From each cluster, we uniformly sample data to train the large model. As shown in Figure~\ref{experiment-fig2}, our solution (\texttt{MLP+Attention}) performs better than \texttt{MLP} because the attention layer considers more semantics. \texttt{no-Hessian} performs the worst because it does not precisely capture the impact of training data instances on the model without the Hessian matrix. 

Also, we conduct experiments to verify the relationship between the Query, Key, Value matrices, which is shown in Figure~\ref{experiment-fig6}.
In this experiment, we compare the Pearson correlation coefficients between the following three methods and the baseline approach, which computes the influence score for the attention layer without any acceleration. (1) \texttt{No-Hessian}($i.e.,$ computing the gradient similarity between  training data and reference data) without considering the Hessian matrix; (2) \texttt{Independent} ($i.e.,$ calculating the Hessian matrices of the query, key, and value layers independently) and (3) \texttt{Ours} ($i.e.,$ calculating the Hessian matrices of the query, key, and value layers as a whole).

\subsection{Ablation Study}
\begin{figure}[h]
\begin{center}
\includegraphics[width=1\textwidth]{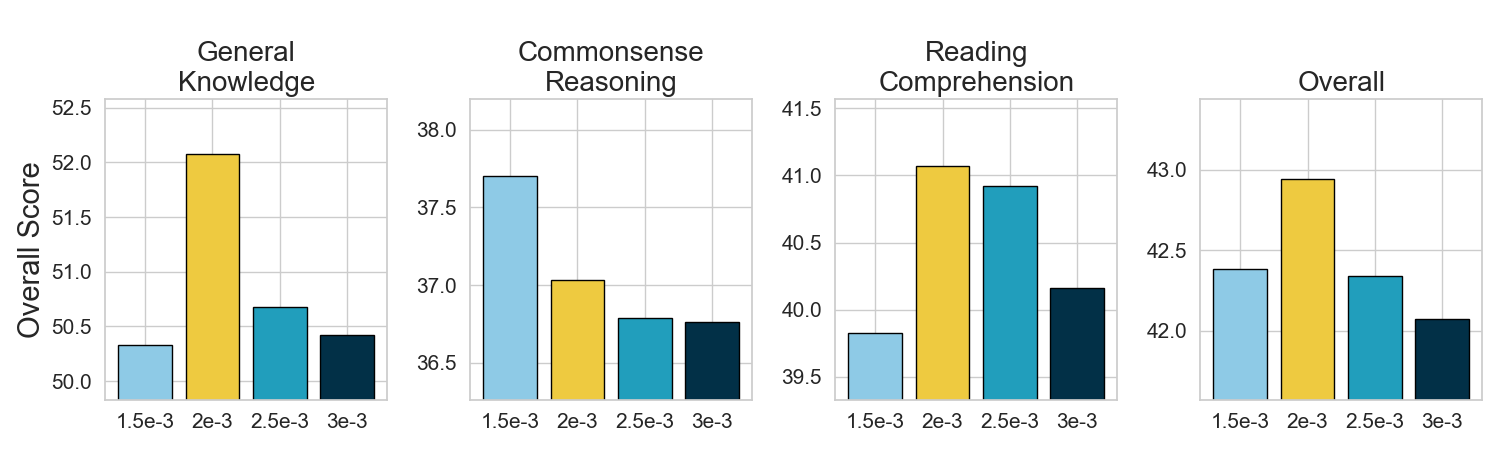}
\end{center}
\caption{$\alpha$ for Quality-Diversity Balance.}
\label{exploration}
\end{figure}

This group of experiments performs ablation studies on the hyperparameters of \texttt{Quad}. Figure~\ref{experiment-fig4} and Figure~\ref{exploration} show the impact of sample threshold and $\alpha$ respectively.


\textbf{Sampling Threshold of Influence ($\tau$).}
 Setting the threshold too high or low will both degrade the model performance.  This is because the selected data instances tend to exist in few clusters with high influence scores, resulting in poor diversity. In contrast, when the threshold is set too low, the sampled instances will be from many clusters with low influence scores, which also degrades the model performance. 

\textbf{$\alpha$ for Quality-Diversity Balance.}
Our approach employs $\alpha$ to balance the diversity and quality in the MAB framework.  
When $\alpha$ is small, we tend to focus on the several clusters with high influence scores without considering diversity much, so the MAB framework is likely to get stuck in a local optimum. 
For example, this results in the model enhancing its performance in specific areas (such as Common Sense Reasoning in Figure~\ref{exploration} when $\alpha = 1.5e-3$), while the performance in other areas ($i.e.,$ General Knowledge and Reading Comprehension) is not good enough. Thus the overall score is not the optimal. 
However, when $\alpha$ is large, the MAB framework focuses too much on diversity without selecting enough high-quality data, which ultimately results in a limited improvement of model performance.

\section{conclusion}
This paper presents \texttt{Quad}, a method designed to balance both the diversity and quality of data in pretraining data selection. \texttt{Quad} employs the influence function to identify data that benefits the model. First, we group the data into clusters and use a subset from each to represent the influence of the entire cluster. Given that influence scores within a cluster display some uncertainty, we view each cluster as an arm in an MAB framework. This method conducts samplings from high-quality clusters, allowing for more precise estimation of their influence scores and meanwhile maintaining the diversity. Moreover, we extend the influence function to attention layers and enhance the calculation efficiency to better measure the impact of data within each cluster on the model.

\bibliography{iclr2025_conference}
\bibliographystyle{iclr2025_conference}

\newpage
\appendix
\section{Appendix}

\begin{table}[h!]
\centering
\caption{Performance Comparison}
\scalebox{0.67}{
\begin{tabular}{lccccccccccccccc}
\toprule
\textbf{Selection Method}& \multicolumn{4}{c}{General Knowledge} & \multicolumn{5}{c}{Commonsense Reasoning} & \multicolumn{3}{c}{Reading Comprehension} & \textbf{Overall} \\
\cmidrule(lr){2-5} \cmidrule(lr){6-10} \cmidrule(lr){11-13} 
& arc-e & arc-c & sciq & \textbf{avg} & logiqa & hellaswag & siqa & winogrande & \textbf{avg} & openbookqa & boolq & \textbf{avg} & \\
\midrule
Random & 50.27 & 20.31 & 80.40 & \textbf{50.33} & 21.20 & 34.11 & 38.49 & 50.99 & \textbf{36.20} & 17.60 & 60.58 & \textbf{39.09} & \textbf{41.55} \\
Semdedup & 51.35 & 20.73 & 80.90 & \textbf{50.99} & 19.05 & 34.56 & 39.30 & 51.54 & \textbf{36.11} & 18.80 & 60.09 & \textbf{39.45} & \textbf{41.81} \\
MATES & 50.00 & 21.25 & 80.10 & \textbf{50.45} & 21.66 & 33.90 & 38.69 & 52.17 & \textbf{36.61} & 19.00 & 60.67 & \textbf{39.84} & \textbf{41.94} \\
PPL & 45.41 & 20.82 & 79.90 & \textbf{48.71} & 20.43 & 35.92 & 39.92 & 54.62 & \textbf{37.72} & 18.80 & 58.35 & \textbf{38.58} & \textbf{41.57} \\
DSIR & 49.28 & 20.14 & 81.70 & \textbf{50.37} & 21.20 & 30.89 & 35.98 & 47.99 & \textbf{34.02} & 16.20 & 61.41 & \textbf{38.81} & \textbf{40.53} \\
Qurating & 52.10 & 23.29 & 79.80 & \textbf{51.56} & 20.43 & 33.57 & 39.05 & 50.67 & \textbf{35.93} & 18.00 & 61.41 & \textbf{39.71} & \textbf{42.04} \\
Quad(ours) & 52.27 & 21.76 & 82.20 & \textbf{52.08} & 22.89 & 34.41 & 38.74 & 52.09 & \textbf{37.03} & 20.00 & 62.14 & \textbf{41.07} & \textbf{42.94} \\
\midrule
top$k$-cluster \\
top-150 & 48.61 & 20.90 & 79.00 & \textbf{49.50} & 23.66 & 34.51 & 39.00 & 51.78 & \textbf{37.24} & 19.20& 61.74 & \textbf{40.47} & \textbf{42.04} \\
top-500 & 51.05 & 21.25 & 79.70 & \textbf{50.67} & 22.73 & 34.40 & 39.20 & 52.41 & \textbf{37.19} & 18.80 & 62.76 & \textbf{40.78} & \textbf{42.48} \\
top-1000 & 49.96 & 20.99 & 80.40 & \textbf{50.45} & 21.97 & 34.00 & 38.74 & 50.2 & \textbf{36.23} & 18.20 & 60.61 & \textbf{39.41} & \textbf{41.67} \\
\bottomrule
\end{tabular}}
\end{table}

\begin{table}[h!]
\centering
\caption{Ablation Study of Threshold $\tau$}
\scalebox{0.7}{
\begin{tabular}{lccccccccccccccc}
\toprule
\textbf{Threshold}& \multicolumn{4}{c}{General Knowledge} & \multicolumn{5}{c}{Commonsense Reasoning} & \multicolumn{3}{c}{Reading Comprehension} & \textbf{Overall} \\
\cmidrule(lr){2-5} \cmidrule(lr){6-10} \cmidrule(lr){11-13} 
& arc-e & arc-c & sciq & \textbf{avg} & logiqa & hellaswag & siqa & winogrande & \textbf{avg} & openbookqa & boolq & \textbf{avg}  & \\
\midrule
0.0015     & 51.26 & 21.16 & 80.20 & \textbf{50.87} & 21.51 & 33.92 & 39.00 & 51.07 & \textbf{36.38} & 19.60 & 61.74 & \textbf{40.67} & \textbf{42.16} \\
0.0020     & 52.23 & 22.27 & 80.70 & \textbf{51.73} & 22.89 & 34.77 & 38.33 & 50.20 & \textbf{36.55} & 19.20 & 61.50 & \textbf{40.35} & \textbf{42.45} \\
0.0025    & 52.27 & 21.76 & 82.20 & \textbf{52.08} & 22.89 & 34.41 & 38.74 & 52.09 & \textbf{37.03} & 20.00 & 62.14 & \textbf{41.07} & \textbf{42.94} \\
0.0030     & 50.25 & 19.62 & 80.80 & \textbf{50.22} & 22.27 & 33.96 & 38.96 & 53.28 & \textbf{37.12} & 20.60 & 59.20 & \textbf{39.90} & \textbf{42.10} \\
\bottomrule
\end{tabular}}
\end{table}

\begin{table}[h!]
\centering
\caption{Ablation Study of $\alpha$}
\scalebox{0.73}{
\begin{tabular}{lccccccccccccccc}
\toprule
\textbf{Alpha} & \multicolumn{4}{c}{General Knowledge} & \multicolumn{5}{c}{Commonsense Reasoning} & \multicolumn{3}{c}{Reading Comprehension} & \textbf{Overall} \\
\cmidrule(lr){2-5} \cmidrule(lr){6-10} \cmidrule(lr){11-13}
& arc-e & arc-c & sciq & \textbf{avg} & logiqa & hellaswag & siqa & winogrande & \textbf{avg} & openbookqa & boolq & \textbf{avg} & \\
\midrule
0.0015 & 50.55 & 20.73 & 79.70 & \textbf{50.33} & 23.35 & 34.60 & 40.58 & 52.25 & \textbf{37.70} & 19.60 & 60.06 & \textbf{39.83} & \textbf{42.38} \\
0.0020 & 52.27 & 21.76 & 82.20 & \textbf{52.08} & 22.89 & 34.41 & 38.74 & 52.09 & \textbf{37.03} & 20.00 & 62.14 & \textbf{41.07} & \textbf{42.94} \\
0.0025 & 51.64 & 22.10 & 78.30 & \textbf{50.68} & 21.81 & 34.76 & 38.02 & 52.57 & \textbf{36.79} & 19.80 & 62.05 & \textbf{40.93} & \textbf{42.34} \\
0.0030 & 50.63 & 21.93 & 78.70 & \textbf{50.42} & 22.12 & 34.64 & 39.15 & 51.14 & \textbf{36.76} & 18.60 & 61.71 & \textbf{40.16} & \textbf{42.07} \\
\bottomrule
\end{tabular}}
\end{table}

\begin{table}[h!]
\centering
\caption{Effectiveness of Influence Calculation}
\scalebox{0.7}{
\begin{tabular}{lccccccccccccccc}
\toprule
\textbf{Method} & \multicolumn{4}{c}{General Knowledge} & \multicolumn{5}{c}{Commonsense Reasoning} & \multicolumn{3}{c}{Reading Comprehension} & \textbf{Overall} \\
\cmidrule(lr){2-5} \cmidrule(lr){6-10} \cmidrule(lr){11-13}
& arc-e & arc-c & sciq & \textbf{avg} & logiqa & hellaswag & siqa & winogrande & \textbf{avg} & openbookqa & boolq & \textbf{avg} & \\
\midrule
Random & 50.27 & 20.31 & 80.40 & \textbf{50.33} & 21.20 & 34.11 & 38.49 & 50.99 & \textbf{36.20} & 17.60 & 60.58 & \textbf{39.09} & \textbf{41.55} \\
No-Hessian & 49.03 & 20.99 & 80.50 & \textbf{50.20} & 22.58 & 33.40 & 38.89 & 52.41 & \textbf{36.82} & 19.20 & 61.50 & \textbf{40.35} & \textbf{42.06} \\
MLP & 50.63 & 21.50 & 78.90 & \textbf{50.34} & 22.89 & 33.32 & 38.74 & 52.57 & \textbf{36.88} & 19.60 & 61.77 & \textbf{40.69} & \textbf{42.21} \\
Ours & 51.05 & 21.25 & 79.70 & \textbf{50.67} & 22.73 & 34.40& 39.20 & 52.41 & \textbf{37.19} & 18.80 & 62.76 & \textbf{40.78} & \textbf{42.48} \\
\bottomrule
\end{tabular}}
\end{table}

\begin{table}[h!]
\centering
\caption{Model Architecture}
\begin{tabular}{ll}
\toprule
{\color[HTML]{1D2129} Hyperparameter} & Value               \\
\midrule
Vocabulary Size                       & 32,000              \\
MLP Ratio                             & 8/3                \\
Hidden Dimension Size                 & 2048                \\
Number of Layers                      & 24                  \\
Number of Attention Heads             & 16                  \\
Number of KV Attention Heads          & 16                  \\
RoPE Base                             & 10,000              \\
Maximum Context Window Length         & 1024                \\
Number of Parameters                  & 1,345,423,360(1.3B) \\
\bottomrule
\end{tabular}
\end{table}


\end{document}